\documentclass[10pt,oneside,onecolumn,a4paper]{article}

\usepackage[utf8]{inputenc}
\usepackage[english]{babel}
\usepackage{bibgerm}
\usepackage{verbatim}
\usepackage{graphicx}
\usepackage{hyperref}
\usepackage[top=3cm, bottom=3cm, left=2.5cm, right=2.5cm]{geometry}

\title{Prosodic entrainment in dialog acts\footnote{{\bf This manuscript is under
  revision. Please contact the authors for information about updates.}}}

\author{Uwe D. Reichel$^1$, Katalin M\'ady$^1$, Jennifer Cole$^2$
  \\$^1$Research Institute for Linguistics, Hungarian Academy of
  Sciences, Budapest, Hungary\\ $^2$Department of Linguistics,
  Northwestern University, Evanston, IL, USA
  \\ uwe.reichel@nytud.mta.hu}

\date{Preprint submitted February 22nd 2018}

\begin{document}

\maketitle

\begin{abstract}
We examined prosodic entrainment in spoken dialogs separately for
several dialog acts in cooperative and competitive games. Entrainment
was measured for intonation features derived from a superpositional
intonation stylization as well as for rhythm features. The found
differences can be related to the cooperative or competitive nature of
the game, as well as to dialog act properties as its intrinsic
authority, supportiveness and distributional characteristics. In
cooperative games dialog acts with a high authority given by knowledge
and with a high frequency showed the most entrainment. The results are
discussed amongst others with respect to the degree of active
entrainment control in cooperative behavior.
\end{abstract}

\section{Introduction}
\label{sec:intro}

In conversation the utterances of speakers become more and more
similar to each other. This phenomenon is called {\em entrainment} or
{\em accommodation} and can be observed at various levels of
linguistic representation.

\subsection{Related work}
Entrainment affects the choice of words \cite{BrennanEP1996,
  Danescu2012, Nenkova2008} and syntactic constructions
\cite{Cleland2003, Gries2005, BraniganC2007}. Entrainment is also
observed in phonetic measures of speaking rate \cite{Levitan2011,
  Levitan2012}, intensity \cite{Levitan2011, Levitan2012}, voice
quality \cite{Levitan2012}, and pitch \cite{Gregory1996, Gregory1997,
  Levitan2011, Babel2012} is reported in data from dialogues and from
speech shadowing experiments. In these studies evidence for
entrainment is derived from acoustic measurements \cite{Levitan2012,
  Babel2012}, word and sentence form analyses \cite{Gries2005}, and
from perceptual similarity ratings \cite{Pardo2006,
  Pardo2013}. Entrainment has also been measured on the basis of
categorical intonation features as derived automatically
\cite{RC_PP16} or by manual labeling \cite{GravanoSLT2014}.

Entrainment is shown to be influenced by the attitudes and the power
relation of the interlocutors, among other factors. Entrainment is
stronger in case of mutual positive attitude of the interlocutors,
than in case of negative attitude \cite{Lee2010}, which is in line
with the predictions of theoretical models such as the Communication
Accommodation Theory (CAT) \cite{Giles1991}. CAT also predicts that
entrainment will be dependent on the dominance relation between
interlocutors. When there is an imbalance in power between two
interlocutors, the one with lower status (or authority/dominance) will
entrain more to the one with higher status \cite{Giles2007}. Empirical
evidence for this claim has been found for talkshow data
\cite{Gregory1996} and for data from the judicial domain
\cite{Benus2012, Danescu2012}, where power hierarchies are well
reflected in the degree of entrainment.

In addition to mutual attitude and power, several other social factors
are correlated with entrainment. As reviewed in \cite{Hirschberg2011}
and \cite{BenusCC2014} entrainment is positively correlated with
perceived social attractiveness \cite{Putnam1984}, mutual likability
\cite{Schweitzer2014}, competence \cite{Street1984}, and
supportiveness \cite{Giles1987}. Remarkably, not only entrainment but
also disentrainment can be positively linked to such social variables
\cite{LoozeSC2014,Perez2016}. \cite{Perez2016} introduced an {\em
  unsigned synchrony measure} not distinguishing between entrainment
and disentrainment but just quantifying their amount in absolute
terms. This unsigned measure was more positively associated with
perceived engagement, encouragement, and the contribution to
successful task completion than a signed measure distinguishing
between entrainment and disentrainment.

The degree of control the speakers have over entrainment is still
under debate. In the CAT framework an active control of entrainment is
assumed in order to negotiate social relations. On the other hand,
\cite{Pickering2004, Pickering2013} and \cite{Chartrand1999} suggest,
that entrainment is rather an automatic mechanism based on a
perception-production link in which the activation of linguistic
patterns increases the likelihood to re-use such patterns. A hybrid
approach is proposed by \cite{Kraljic2008, Lewandowski2012,
  Schweitzer2014}, in that accommodation is partially automatic but
also actively controlled to adapt to within- and inter-speaker
influences. Simulations of this hybrid mechanism have been computed
\cite{Lewandowski2017} in an exemplar-theoretic framework
\cite{Nosofsky1986, Johnson1997}.

Regardless of the degree of active control over entrainment, its
benefit is well documented: entrainment has been shown to increase the
success of conversation in terms of low inter-turn latencies, a
reduced number of interruptions \cite{Levitan2012, Nenkova2008}, and
for objective task success measures \cite{Reitter2014}. Related to the
success of conversation \cite{RC_PP16} found more entrainment in
cooperative than in competitive dialogs for prosodic event
sequences. These findings on social variables and conversation success
are again in line with the claim from CAT that entrainment enhances
social approval and communication efficiency.

In the empirical work mentioned above entrainment was measured by
comparing the properties of more and less closely related units,
i.e. adjacent vs. non-adjacent speech segments, or segments from the
same vs. from different dialogs (local and global entrainment,
respectively; see \cite{LoozeSC2014} for a systematic overview). For
spoken dialogs these units are most commonly turns
(e.g. \cite{Levitan2011}) or stretches of speech in a fixed time frame
(e.g. the TAMA approach by \cite{Kousidis2008}). But so far very
little work has been done for units that are defined with respect to
their {\em function} in a dialog, namely dialog acts. For this reason
it is not well known yet to what extent entrainment depends on
functional dialog units. One work that attempts to relate entrainment
to dialog function is \cite{Benus2014}, which examined prosodic
entrainment in Slovak for the discourse particle {\em no} with respect
to usage statistics and several acoustic parameters. {\em No} serves
to signal affirmation, backchannel, and mild disagreement. It was
shown that for parts of the underlying data that frequency entrainment
was absent for {\em no} in general but present for {\em no} with one
of its discourse functions. In other work, \cite{Mizukami2016} found a
higher amount of lexical entrainment for dialog acts with little
informational content as greeting, closing, backchannel, and
agreement. Less entrainment was found for dialog acts expressing
opinions as apologies and action directives.

\subsection{Goals of this study}
\paragraph{Hypotheses}
This study aims to contribute to this yet understudied aspect of
entrainment as it relates to dialog acts. Our focus is on prosodic
entrainment, which we investigate in speech from an interactive game
task where participants play under cooperative and competitive game
conditions. We examine acoustic evidence of prosodic entrainment as a
function of the dialog act of the utterance in cooperative and
competitive play. Dialog acts are differentiated along the social
dimensions of authority and supportiveness, by frequency, and by local
predictability in order to test these influences on entrainment
behavior on the dialog act level. Based on the findings from the cited
prior work on the effects on entrainment of interlocutor attitudes and
status relations, and work showing the benefit of entrainment for task
success we formulate the following hypotheses:

\begin{itemize}
\item[H1] There is more entrainment in intrinsically low than in high
  authority dialog acts.
\item[H2] There is more entrainment in supportive dialog acts than in
  dialog acts that are neutral or negative in providing support for
  the interlocutor.
\end{itemize}

Low and high authority dialog acts both cause local authority
imbalances in that they decrease or increase the speaker's authority
relative to the interlocutor. Hypothesis 1 thus serves to examine
whether this local imbalance has a similar effect on entrainment as
general authority imbalances in the judicial data of \cite{Benus2012,
  Danescu2012}.

The effect of supportive dialog acts in boosting entrainment is
restrictedly expected for speech produced in cooperative game play,
where interlocutors must work together towards a shared
goal. Supportive dialog acts are predicted to be rare in competitive
game play, and accordingly we focus on cooperative play for testing
Hypothesis 2.

Additional hypotheses (H3, H4) relate to the frequency and
predictability of a dialog act. Entrainment is positively correlated
with task success, and is expected to be optimized in conditions of
cooperative interaction. In cooperative game play, optimization can be
achieved when entrainment is concentrated in the most frequent dialog
acts, where the expected benefit of smooth turn transitions can be
maximized. Optimization can also be achieved through the selective use
of disentrainment with dialog acts that are locally unpredictable. In
such cases disentrainment disrupts common prosodic patterns and in
this way may serve to attract the interlocutor's attention.

\begin{itemize}
\item[H3] Entrainment will be greater in frequent dialog acts than in
  less frequent ones.
\item[H4] Disentrainment will be more frequent in locally
  unpredictable dialog acts.
\end{itemize}

H3 extends the above mentioned studies on the positive impact of
entrainment on smooth turn transitions \cite{Levitan2012,
  Nenkova2008}. H4 describes a possible concrete case of a cooperative
disentrainment behavior which is more generally suggested by studies
such as \cite{Perez2016} referred to above.

\paragraph{Implication}
If entrainment differs among dialog acts and between cooperative and
competitive settings, this can be taken as an indication that it
cannot only be an automatic process in terms of a perception-action
loop but is also at least partially actively controlled.

\paragraph{Prosodic stylization}
Our approach extends the prosodic feature set used in prior studies of
prosodic entrainment. Prosodic analyses in prior work (cited above)
are restricted to simple acoustic measures like the mean or maximum
value of fundamental frequency (f0) \cite{Babel2012, Levitan2011,
  Levitan2012}, and its variance \cite{Gregory1996}. In the study
presented here, we add features derived from a parametric
superpositional intonation stylization, that allow for the comparison
of complex and temporally dynamic pitch patterns in different prosodic
domains.

\section{Data}
Our analysis is based on a subset of the speech data from the Illinois
Game Corpus \cite{Page,RPNC_IS2015} that is comprised of Tangram game
dialogs by American English speakers in cooperative and competitive
settings. The tangram is a puzzle consisting of seven pieces that can
be combined to form shapes that resemble various common objects, such
as a boat, house or person.  Both dialog partners were separately
presented with Tangram silhouettes that were hidden from the view of
the other partner. The task was to decide whether the silhouettes were
the same or different by verbally describing them to each other.  In
the cooperative setting the partners solved this common goal in a
joint effort. In the competitive setting, the partners were required
to solve this task competitively, and the one who solved the puzzle
first over the most number of trials was awarded a candy prize as the
winner. Undergraduate students (ages 18--29) from the University of
Illinois, all native monolingual speakers of American English, were
recruited as paid participants in this study. Twelve pairs of
participants took part in the experiment. They were prompted to engage
in free conversation for a few minutes after which they played the
Tangram game together, first playing cooperatively and then
competitively, with different Tangram silhouette images in each
condition. Participants were seated in chairs facing one another, with
no intervening table and with the printed Tangram silhouettes
positioned off to the side, facing each participant.  Audio and video
recordings were made on separate channels for each
participant. Participants provided written consent for the use of
these recordings in research. For the current study a dialog-act
annotated subset of 16 dialogs (10 cooperative, 6 competitive) by 11
interlocutor pairs was used, of which eight were Female-Female pairs
and three were Male-Female pairs. Mean dialog duration amounts to 7
minutes 40 seconds.  The used part of the corpus was manually
dialog-act segmented and annotated using the tag set of
\cite{Carletta1996, Carletta1997}, which is described in more detail
in section \ref{sec:da}. Additional tags e.g. for comments and offtalk
(see \cite{RPNC_IS2015}) not belonging to the original tag set were
ignored for the current study, so that the examined data consists of
4011 dialog act segments.

\section{Dialog acts}
\label{sec:da}

\subsection{Inventory}

The applied tagset was developed by \cite{Carletta1996, Carletta1997}
in order to describe conversational moves, i.e. initiations and
responses with certain discourse purposes. The complete label set is
shown in Table \ref{tab:da} and is briefly described following the
guidelines given in \cite{Carletta1997} and illustrated by some
examples from our corpus. Dialog acts were labeled in parallel by two
annotators working with the text transcriptions alone (no
audio). Mismatches were subsequently resolved by discussion between
these two.

\begin{itemize}
\item {\bf Acknowledgment AC.} Listener feedback e.g. to signal
  accordance or acceptance. Examples: {\em Ok., Yeah., Like you're
    thinking.}
\item {\bf Alignment AL.} Checks the attention, agreement, or
  readiness of the interlocutor. Examples: {\em Or do you want more?,
    Ok ready?}
\item {\bf Check CH.} Requests the interlocutor to confirm an
  information. Examples: {\em Like this?, Ok so so it has an open
    door?}
\item {\bf Clarify CL.} A reply that includes additional information
  which was not explicitly asked for. Example: {\em it's like the
    house and then it's like right next to it there's like a horse
    stable (answer to ``Does it have like a little like a little thing
    on the side on the right)., It's more like it's inwards towards
    the horse's head (answer to ``ok so if he was riding the horse
    does it look like his chest would be sticking out then?'').}
\item {\bf Explain EX.} Providing information not directly elicited by
  the partner (thus no reply) Examples: {\em It looks like a more
    geometric batman symbol., The bottom looks like a person it looks
    like a person in a boat.}
\item {\bf Instruct IN.} Commands the interlocutor to perform an
  action. Examples: {\em Hold on!, Look at me!}
\item {\bf Question-W QW.} Wh-question. Examples: {\em So what what
  image do you think we have?, Does yours have one or two legs?}
\item {\bf Question-YN QY.} Yes-no question. Examples: {\em Does it
  have a door?, Are the arms like this kinda arms?}
\item {\bf Ready RE.} Indicating readiness to go on (here as opposed
  to \cite{Carletta1997} not restricted to game initial
  position). Examples: {\em Ok!, Alright!}
\item {\bf Reply-No RN.} No-reply. Examples: {\em No., Head head's not
  down (answer to ``Head down?'').}
\item {\bf Reply-W RW.} A reply conveying more information than ``yes''
  or ``no'' but not more than what was asked. Examples: {\em Like a
    diamond face (answer to ``And is it like a full face? Like a like
    a this face?''), Two legs (answer to ``Does yours have one or two
    legs?'').}
\item {\bf Reply-Y RY.} Yes-reply. Examples: {\em Yes., You can go first
  (answer to ``Can I go first?'').}
\end{itemize}

\subsection{Grouping}

In order to test the hypotheses formulated in section \ref{sec:intro}
we subdivided the dialog acts along 4 dimensions: authority,
supportiveness, frequency, and local predictability. These dimensions
are summarized in Table \ref{tab:da}, with counts for each dimension
in Table \ref{tab:grp_cnt} and detailed explanations in the following
paragraphs.

\paragraph{Authority}
Following \cite{Bochenski1974, George1985} authority is given by
knowledge (epistemic authority) or by a superior position which
enables a person to give orders (deontic/executive authority). We thus
clustered all dialog acts fulfilling one of these two conditions to
the high authority group, and all others to the low authority group.
Examples for high-authority dialog acts are EX and CL fulfilling the
knowledge condition, and IN to influence the interlocutor. Low
authority dialog acts are those that are usually neutral with respect
to dominance (AC, AL, RE), or reflect a dependency of the speaker on
the cooperation of the interlocutor, which generally holds for
non-executive request (AL, CH, QW, QY). Alignment AL in principle
could also express a dominance relation, but this was not observed in
our data, so we assigned AL to the low-authority group.

\paragraph{Supportiveness}
Dialog acts further can be subdivided into two groups according to the
degree of their supportiveness. We consider a dialog act to be
supportive if it helps the interlocutor to get to a common ground
\cite{Clark1989}, i.e. if it provides information (EX, CL, RY, RN, RW)
or serves to ensure that given information was understood (AL, AC).

\paragraph{Frequency}
The frequency distinction was simply derived by calculating the
probability of each dialog act in our corpus and setting the median
probability value as the boundary dividing the high- and low-frequency
dialog acts. We used probabilities instead of raw frequencies to allow
for later cross-corpus comparisons.

\paragraph{Local predictability}
Local predictability does not provide an overall dialog act
categorization but classifies each dialog act in each context it
occurs. As an approximation this local context is given by the
preceding dialog act in the dialog, so that local predictability can
be measured in terms of dialog act bigram probabilities
$P(da_i|da_{i-1})$. This is the conditional probability of the dialog
act at position $i$ given the preceding dialog act. Both for the
unigram probabilities above and for the bigram probabilities maximum
likelihood estimates were used. Again, the median value of the bigram
probabilities was taken to divide the dialog acts in context into a
high and a low predictability group.

\begin{table}
\begin{center}
\begin{tabular}{lllll}
\hline
id & dialog act & authority & support & frequency \\
\hline
AC & Acknowledgment &        low &      yes &       high \\
AL & Alignment &       low &      yes &        low \\
CH & Check &       low &       no &       high  \\
CL & Clarify &      high &      yes &        low  \\
EX & Explain &      high &      yes &       high  \\
IN & Instruct &      high &       no &        low \\
QW & Question-W    &   low &       no &        low \\
QY & Question-YN    &   low &       no &       high \\
RE & Ready &      low &       no &       high \\
RN & Reply-No &     high &      yes &        low \\
RW & Reply-W &     high &      yes &        low \\
RY & Reply-Y &      high &      yes &       high \\
\hline
\end{tabular}
\caption{Dialog acts and their grouping.}
\label{tab:da}
\end{center}
\end{table}

\begin{table}
\begin{center}
\begin{tabular}{ll|ll|ll|ll}
\hline
\multicolumn{2}{c|}{authority} & \multicolumn{2}{c|}{support} & \multicolumn{2}{c|}{frequency} & \multicolumn{2}{c}{predictability} \\
high & low & yes & no & high & low & high & low \\
\hline
1982 & 2029 & 2651 & 1360 & 3558 & 453 & 3813 & 198
\end{tabular}
\caption{Number of dialog act segments for each dimension's levels.}
\label{tab:grp_cnt}
\end{center}
\end{table}

\section{Prosodic analyses}

Our goal is to assess prosodic entrainment in pitch, loudness and
tempo, considering general global measures (maximum, median, standard
deviation) that hold of dialog acts, and local measures that hold of
prosodic phrases and pitch accent domains within the dialog act. We
also assess entrainment in the temporally dynamic patterns of f0
across prosodic phrases and accent domains. We focus the analysis of
local measures on the initial and final prosodic phrases, and the
initial and final pitch accents in a dialog act. These initial and
final regions demarcate the prosodic structuring of the act and are
the locations where critical and obligatory intonational features are
specified (see \cite{Ladd2008} for an overview). To extract these
acoustic measures we use automated methods only, which enables
replication and comparison of findings among different speech
datasets.

\subsection{Preprocessing}

\paragraph{Transcription and Alignment}

Audio files for each dialog were manually segmented into chunks and
orthographically transcribed.  The text within each chunk was then
automatically aligned to the signal using the multilingual WEBMAUS
webservice \cite{krsCSL2017} with the parameter setting for American
English. For the sake of subsequent prosodic structure inference, word
stress was added to the phonemic transcriptions by the help of the
Balloon Grapheme-Phoneme Converter \cite{ReichelIS2012} also available
as a webservice \cite{krsCSL2017}.

\paragraph{F0 and energy}

F0 was extracted by autocorrelation (PRAAT 6.0.35 \cite{Boersma1999},
sample rate 100 Hz). Voiceless utterance parts and f0 outliers were
bridged by linear interpolation. The contour was then smoothed by
Savitzky-Golay filtering \cite{Savitzky1964} using third order
polynomials in 5 sample windows and transformed to semitones relative
to a base value. This base value was set to the f0 median below the
5th percentile of an utterance and served to normalize f0 with respect
to its overall level. Energy in terms of root mean squared deviation
was calculated with the same sample rate as f0 in Hamming windows of
50 ms length.

\paragraph{Prosodic structure}

The following prosodically relevant time points were extracted
automatically within each chunk using the open source CoPaSul toolkit
\cite{ReichelArxiv2016,copasulGithub}: syllable nuclei, prosodic
phrase boundaries, and pitch-accented syllables as described in detail
in \cite{reichelESSV2017}. Syllable nucleus assignment follows the
procedure introduced in \cite{Pfitzinger1996} to a large extent. An
analysis window $w_a$ and a reference window $w_r$ with the same time
midpoint were moved along the band-pass filtered signal in 50 ms
steps.  For syllable nucleus assignment the energy is required to be
higher in $w_a$ than in $w_r$ by a certain factor, and additionally
had to surpass a threshold relative to the maximum energy of the
recording.

Phrase boundaries were detected automatically by means of a
bootstrapped nearest centroid classifier. From pitch register
discontinuity features derived for each right-edge word boundary
\cite{ReichelMadyIS2014} and from vowel length z-scores two centroids
for phrase-final and non-final word boundaries were bootstrapped based
on two simplifying assumptions: (1) each pause is preceded by a
boundary, and (2) since prosodic phrases have a minimum length, in the
vicinity of pauses in both directions there are no further
boundaries. The minimum length was set to 1 second. From this initial
clustering feature weights were calculated from the mean cluster
profile derived separately for each feature.  The remaining word
boundaries were then classified as phrase boundaries or not phrase
boundaries, based on their weighted Euclidean distances to the two
centroids.

Pitch accents were detected in an analogous fashion, using local pitch
shape and energy features within the word-stressed syllables and by
two different simplifying assumptions for cluster initialization: (1)
all words longer than a threshold $t_a$ in seconds are likely to be
content words that contain a high amount of information and are thus
taken as ``accented'' representatives, and (2) all words shorter than
a threshold $t_{na}$ are likely to be function words with a low amount
of lexical information and are thus taken as ``no accent''
representatives. $t_a$ and $t_{na}$ were set to 0.5 and 0.1 sec,
respectively, thus to rather extreme word length values in order to
increase the precision of the initial candidate selection.

In \cite{reichelESSV2017} this procedure was optimized with respect to
F1 scores on spontaneous speech data and yielded F1 values of 0.61 and
0.63 for boundary and accent detection, respectively. This indicates
rather moderate precision and recall values, which is a trade-off to
the advantages of the automated processing described above. However,
the value ranges of the chosen features -- pitch discontinuity for
boundaries, pitch shape and energy for accents -- are split by the
clustering such that boundaries are placed at high discontinuities,
and syllables with salient pitch and energy movements are identified
for further accent analyses. In other words, the automated procedure
is tuned to avoid false positives, at the cost of not detecting
boundaries and accents that have lower acoustic salience.

\subsection{Feature extraction}
\begin{figure}
  \begin{center}
    \includegraphics[width=12cm]{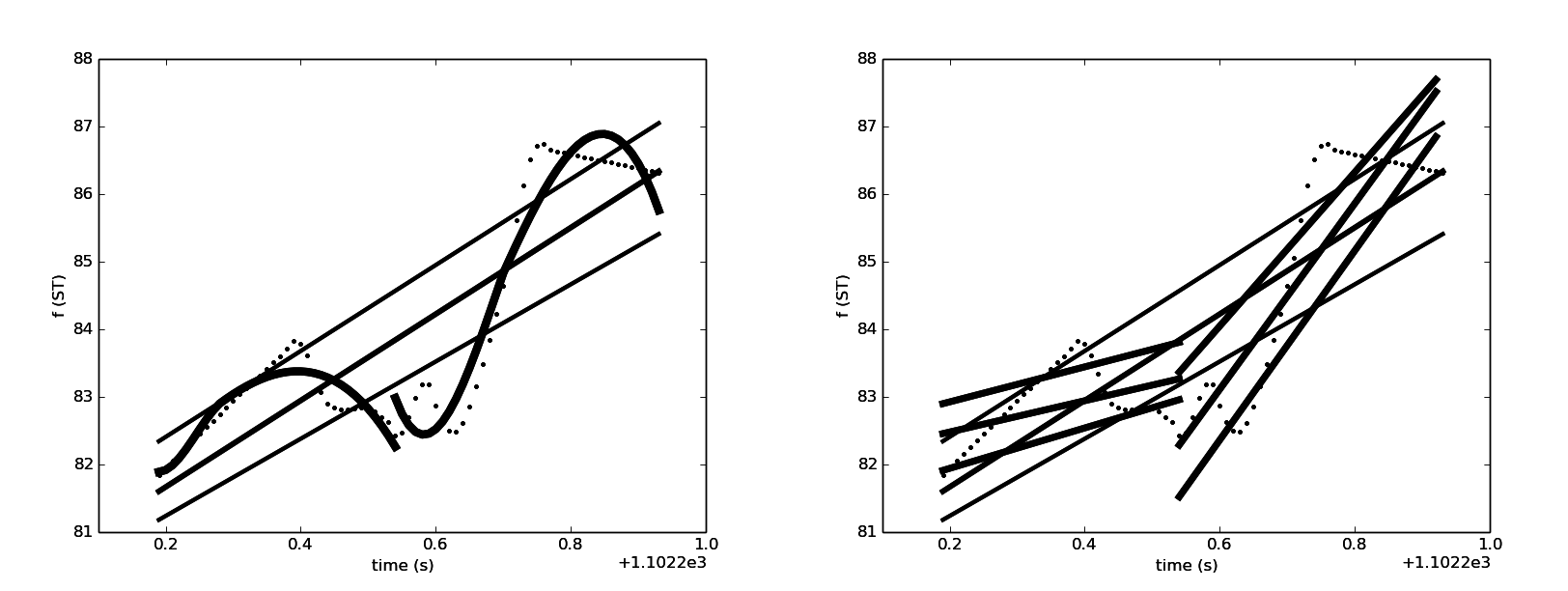}
    \caption{Superpositional f0 stylization within the CoPaSul
      framework. On the prosodic phrase level a base, mid- and topline
      (solid) are fitted to the f0 contour (dotted) for register
      stylization. Level is represented by the midline, and range by a
      regression line fitted to the pointwise distance between base
      and topline. On the local pitch event level comprising accents
      and boundary tones the f0 shape is represented by a third-order
      polynomial, one for each of the two events (left). The f0
      Gestalt properties, i.e. its register deviation from the
      phrase-level register is quantified by generating a local
      register representation the same way as for the phrase level
      (right) and by calculating the root mean squared deviations
      between the midlines and the range regression lines.}
    \label{fig:superpos}
  \end{center}
\end{figure}

In addition to the general f0 and energy features mentioned above, we
derived features related to pitch register and the local pitch event
from the contour-based, parametric, and superpositional CoPaSul
stylization framework \cite{ReichelCSL2014}, which represents f0 as a
superposition of a global register and a local pitch accent
component. This stylization is presented in Figure
\ref{fig:superpos}. Rhythmic features were also extracted, as
described below. All features introduced here can again be extracted
by means of the open source CoPaSul prosody analysis software
\cite{ReichelArxiv2016, copasulGithub}.

All features are listed in Table \ref{tab:feat} along with the feature
set name they belong to and a short description. A more detailed
description is given in the subsequent sections.

\begin{center}
  \begin{table}
    \begin{tabular}{lll}
      \hline
      Feature set & Feature & Description \\
      \hline
      GEN & max & energy maximum in dialog act\\
      GEN & med & energy median in dialog act\\
      GEN & sd & energy standard deviation in dialog act\\
      \hline
      GF0 & max & f0 maximum in dialog act\\
      GF0 & med & f0 median in dialog act\\
      GF0 & sd & f0 standard deviation in dialog act\\
      \hline
      IP & rng.c0.F/L & f0 range intercept of first/last phrase \\
      IP & rng.c1.F/L & f0 range slope of first/last phrase \\
      IP & lev.c0.F/L & f0 level intercept of first/last phrase \\
      IP & lev.c1.F/L & f0 level slope of first/last phrase \\
      \hline
      ACC & c0-3.F/L & polynomial coef of the first/last pitch accent \\
      ACC & rng.c0.F/L & f0 range intercept of first/last pitch accent \\
      ACC & rng.c1.F/L & f0 range slope of first/last pitch accent \\
      ACC & lev.c0.F/L & f0 level intercept of first/last pitch accent \\
      ACC & lev.c1.F/L & f0 level slope of first/last pitch accent \\
      ACC & gst.lev.F/L & f0 level deviation of first/last pitch accent \\
      ACC & gst.rng.F/L & f0 range deviation of first/last pitch accent \\
      \hline
      RHY & syl.rate & mean syllable rate \\
      RHY & syl.prop.en & syllable influence on energy contour \\
      RHY & syl.prop.f0 & syllable influence on f0 contour \\
    \end{tabular}
    \caption{Description of prosodic features grouped by feature sets.}
    \label{tab:feat}
  \end{table}
\end{center}

\subsection{General f0 and energy features}

For the feature sets {\em GEN} and {\em GF0} for general energy and f0
characteristics within each dialog act we calculated the median, the
maximum, and the standard deviation of the f0 and the energy contour,
respectively.

\subsection{Prosodic phrase characteristics}
\label{sec:glob}

The {\em IP} feature set describes f0 register characteristics of the
intonational phrase. According to \cite{Rietveld2003} f0 register in
the prosodic phrase domain can be represented in terms of the f0 range
between high and low pitch targets, and the f0 mean level within this
span. To capture both register aspects and their change over time,
within each prosodic phrase we fitted a base-, a mid, and a topline by
means of linear regressions as shown in Figure
\ref{fig:superpos}. This line fitting procedure works as follows: A
window of length 50 ms is shifted along the f0 contour with a step
size of 10 ms. Within each window the f0 median is calculated (1) of
the values below the 10th percentile for the baseline, (2) of the
values above the 90th percentile for the topline, and (3) of all
values for the midline. This gives three sequences of medians, one
each for the base-, the \mbox{mid-,} and the topline, respectively. These
lines are subsequently derived by linear regressions, with time
normalized to the range from 0 to 1. As described in further detail in
\cite{ReichelMadyIS2014} this stylization is less affected by local
events such as pitch accents and boundary tones and does not need to
rely on error-prone detection of local maxima and minima. Based on
this stylization the midline is taken as a representation of pitch
level.  For pitch range we fitted a further regression line through
the pointwise distances between the topline and the baseline. A
negative slope thus indicates convergence of top- and baseline,
whereas a positive slope indicates divergence.

From this register level and range representation we extracted the
following features for the first and for the (occasionally identical)
last prosodic phrase in (or overlapping with) a dialog act: intercept
and slope of the midline, and intercept and slope of the range
regression line. That gives eight features subsumed to the {\em IP}
feature set.

\subsection{Pitch accent characteristics}
\label{sec:loc}

We next normalized each f0 value to the corresponding local range with
the two reference points on the base- and topline (cf section
\ref{sec:glob}) set to 0 and 1, respectively. By this normalization f0
values between base- and topline range from 0 to 1, f0 values below
the baseline are smaller than 0, and values above the topline are
greater than 1. We fitted third-order polynomials to this f0 contour
residual around the syllable nuclei associated with the first and the
last local pitch event (accent or boundary tone) in a dialog act. The
stylization window of length 300 ms was placed symmetrically on the
syllable nucleus, and time was normalized to the range from $-1$ to
1. This window length of approximately 1.5 syllables was chosen to
capture the f0 contour on the accented syllable in some local context.

The coefficients of the fitted polynomials represent different aspects
of local f0 shapes. Given the polynomial $\sum_{i=0}^3 s_i\cdot t^i$
for the normalized time variable $t$, the coefficient $s_0$ is related
to the local f0 level relative to the local range.  $s_1$ and $s_3$
are related to the local f0 trend (rising or falling) and to peak
alignment. $s_2$ determines the peak curvature (convex or concave) and
its acuity.  Next to the polynomial coefficients we measured local
register values by re-applying the stylization introduced in section
\ref{sec:glob} within the analysis window around the pitch accent.

Finally, pitch accent Gestalt was measured in terms of local register
deviation from the corresponding stretch of global register. This was
simply done by calculating the root mean squared deviation (RMSD)
between the pitch accent midline and the corresponding part of the
phrase midline. For the accent and phrase range regression lines the
same procedure was used.

From these stylizations the feature set {\em ACC} emerges for the
first and for the last local pitch event in a dialog act. It contains
(1) the polynomial coefficients describing the local f0 shape, (2) the
intercept and slope coefficients for the mid- and the range regression
line describing the local register, and (3) the local level and range
deviation from the underlying phrase in terms of the RMSD between the
accent- and phrase-level regression lines.

\subsection{Rhythm features}

The {\em RHY} feature set captures at the level of the dialog act
properties traditionally termed as ''rhythmic'', including syllable
rate (number of detected syllable nuclei per second) and the influence
of the syllable level of the prosodic hierarchy on the energy and f0
contours. Influence means, to what extent the syllable oscillator
determines the shape of these contours. This influence manifests
itself in regular fluctuations at the syllable rate. To quantify the
syllabic influence on any of these contours we performed a discrete
cosine transform (DCT) on this contour as in
\cite{HeinrichJASA2014}. We then calculated the syllable influence $w$
as the relative weight of the coefficients around the syllable rate
$r$ ($+/- 1$ Hz to account for syllable rate fluctuations) within all
coefficients below 10 Hz as follows:

\begin{eqnarray}
w & = & \frac{\sum_{c: r-1 \leq f(c) \leq r+1 \textrm{Hz}} |c|}{\sum_{c: f(c)\leq 10 \textrm{Hz}} |c|} \nonumber
\end{eqnarray}

\begin{figure}
  \begin{center}
    \includegraphics[width=10cm]{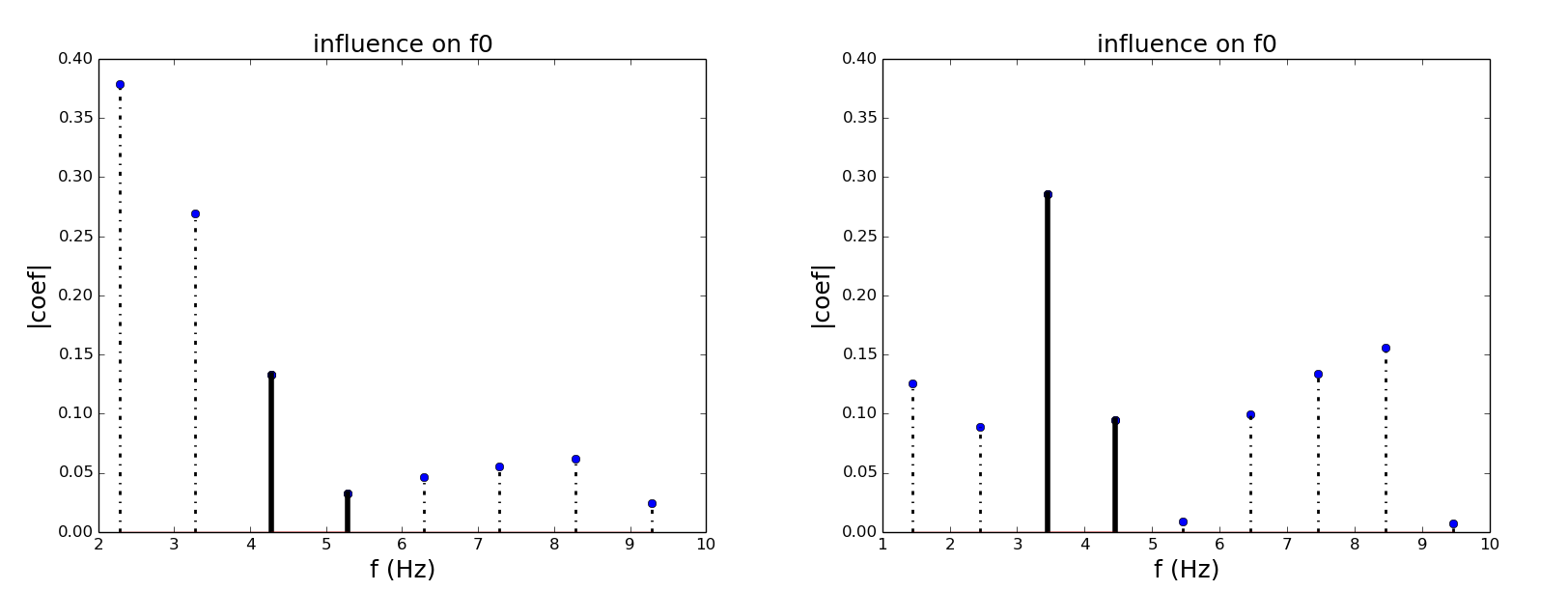}
    \caption{Rhythm features: Quantifying the influence of syllable
      rate on the f0 contour (analogously for the energy contour). For
      this purpose a discrete cosine transform (DCT) is applied to the
      contour. The absolute amplitudes of the coefficients around the
      syllable rate are summed and divided by the summed absolute
      amplitudes of all coefficients below 10 Hz. This gives the
      proportional influence of the syllable on the contour. In the
      left case the syllable oscillator (syllable rate is 4.8 Hz) has
      a relatively low impact on the f0 contour whereas on the right
      the impact of the 4 Hz syllable oscillator is relatively
      high. Conversely, the impact of the lower-frequency pitch accent
      oscillator in the 2 Hz region is high in the left case and low
      on the right.}
    \label{fig:rhy}
  \end{center}
\end{figure}

The higher $w$ the higher thus the relative influence of the syllable
rate on the contour. Furthermore, a high relative syllable influence
implies a lower impact of other macroprosodic oscillators as pitch
accents and vice versa, so that $w$ also can be regarded as an inverse
measure of pitch accent influence. This procedure which is shown in
Figure \ref{fig:rhy} was first used to quantify the impact of hand
stroke rate on the energy contour in counting out rhymes
\cite{FR_PP16}. The upper cutoff of 10 Hz goes back to the reasoning
that contour modulations above 10 Hz do not occur due to macroprosodic
events as accents or syllables, but amongst others due to
microprosodic effects.

\section{Entrainment measurement}

\subsection{Method}
\label{sec:met}

In this study we focus on global entrainment, i.e. we compare
identical dialog act pairs within a dialog with pairings between
speakers not engaged in any common game conversation. The
within-dialog sample was generated as follows: for each dialog act of
speaker A we randomly picked one dialog act of the same kind uttered
by speaker B from the preceding course of the dialog, if
available. For the across-dialog sample we randomly paired the dialog
act of speaker A with one dialog act of the same kind uttered by an
unrelated speaker C from another dialog. Being unrelated further
implies that A and C did not engage in any common conversation in this
corpus.

\begin{figure}
  \begin{center}
    \includegraphics[width=10cm]{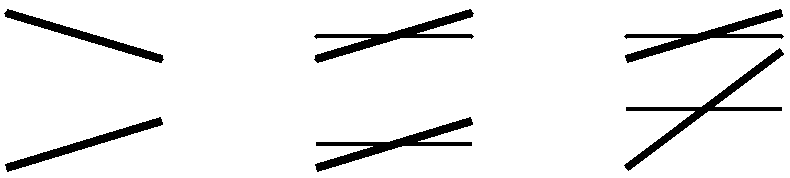}
    \caption{Entrainment in terms of convergence (left), synchrony
      (mid), and both (right).}
    \label{fig:meas}
  \end{center}
\end{figure}

As pointed out in \cite{Levitan2011, LoozeSC2014} accommodation can be
expressed, amongst others, in terms of convergence or synchrony. As
visualized in Figure \ref{fig:meas} convergence means that feature
values become more similar. Convergence-related distance is trivially
represented by the absolute distance of the feature value pair, the
lower the distance, the higher the convergence. Synchrony means that
feature values vary in parallel. \cite{LoozeSC2014} proposes to
calculate correlations over a sequence of segment pairs. Here we
choose a more straight-forward approach operating on a single dialog
act segment pair only. We simply subtract the respective speakers'
mean values from the feature values before calculating the absolute
distance. Synchrony-related distance is thus low, if the speakers
realize a feature either both above or below their respective
means. By that we derive for each feature and each dialog act segment
pair one convergence- and one synchrony-related distance value.
Clearly, and as depicted in Figure \ref{fig:meas}, the terms
"convergence" and "synchrony" describe patterns of change over time
for two signals. We have operationalized these notions in terms of
static measures here, but in the remainder of this paper, for the sake
of readability, we abbreviate {\em convergence-} and {\em
  synchrony-related distance} by convergence and synchrony,
respectively.

\subsection{Entrainment by dialog act}

For an initial harvesting of the data separately for cooperative and
competitive dialogs we statistically compared the within- and
across-dialog differences by two-sided t-tests for independent samples
for each of the 12 dialog act types, for each of the 5 feature sets,
and for the 2 distance measures. The significance level was set to
$0.05$.

\subsection{Entrainment by dialog act grouping}

In order to test the effects of the dialog act groupings on
entrainment we pooled all data across the 5 feature sets and the 2
entrainment measures in the following way: within each dialog act
segment we obtained for each single feature and each of the
entrainment measures (convergence and synchrony) 2 values as described
in section \ref{sec:met}, a within the same dialog distance, $d_s$,
and an across different dialogs distance, $d_d$. We then simply
subtracted $d_d$ from $d_s$ to obtain the distance delta $d$. $d$
values well below 0 thus reflect a greater within dialog similarity
and indicate an entrainment tendency, whereas values well above 0
indicate a disentrainment tendency. We tested the effect of the dialog
act grouping on entrainment by two linear mixed effects models, one
with the fixed effects authority, support, and dialog condition
(cooperative vs. competitive), and the other with the fixed effects
frequency, local predictability, and again dialog condition. In both
models the dependent variable is given by $d$, and the speaker
uttering the dialog act from which $d$ is calculated is taken as a
random effect. In both tests a random slope model was calculated for
the speaker Id and the fixed effects. In case of significant
interactions the models were re-applied on the respective subsets.
For the linear mixed effects models we used the R function {\em lmer}
from the package {\em lme4} \cite{lme4} and for $p$-value assignment
the R function {\em Anova} from the package {\em car} \cite{car}.

\section{Results}

\subsection{Entrainment by dialog act}

\paragraph{Profiles}

\begin{figure}
  \begin{center}
    \includegraphics[width=12cm]{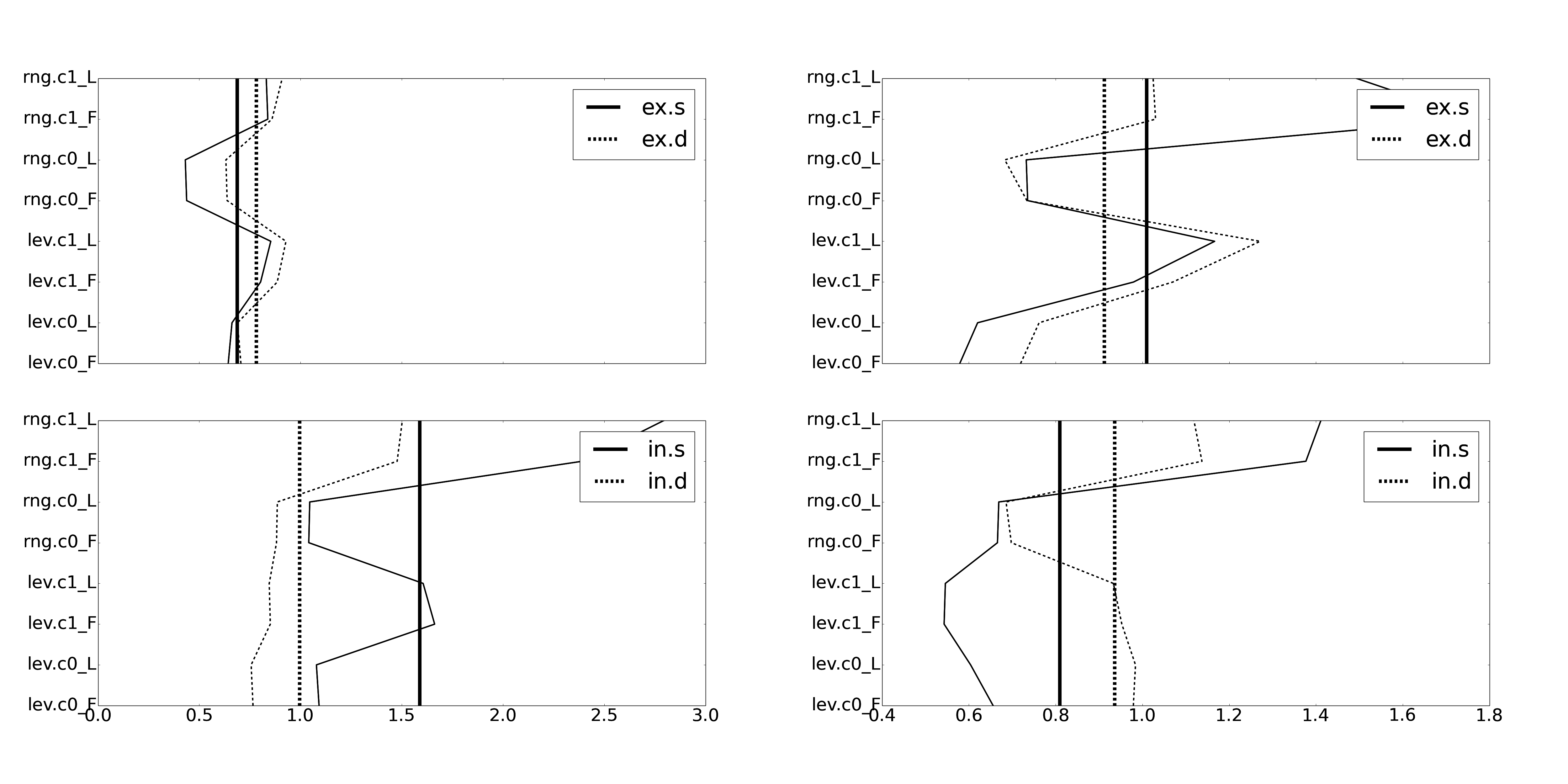}
    \caption{Entrainment profiles for the dialog acts EX (top row) and
      IN (bottom row) and the feature set IP in cooperative (left) and
      competitive (right) dialogs. The solid, vertical (straight)
      lines give the mean distances in terms of convergence across
      randomly picked dialog instances in the same dialog (\_s). The
      dashed vertical (straight) lines represent mean distances across
      randomly picked unrelated speaker pairs (\_d). The means are
      calculated over all features in the set IP. Values for each
      feature on the y-axis shown in thin (jagged) solid lines (\_s),
      and thin dashed lines (\_d). An entrainment tendency is
      indicated by a solid vertical line left of its dashed
      counterpart. For disentrainment the solid line is right of the
      dashed one.}
    \label{fig:profil}
  \end{center}
\end{figure}

Figure \ref{fig:profil} shows entrainment profiles for two dialog acts
EX ({\em explain}) and IN ({\em instruct}) in cooperative and
competitive dialogs for the feature set IP and the convergence
distance measure. The solid vertical lines give the mean within dialog
distances $d_s$ of the features in the set IP, and the dashed lines
the mean across dialog distances $d_d$. A solid line left of its
dashed counterpart indicates entrainment, and the opposite order
indicates disentrainment. It can be seen that the entrainment profiles
of EX and IN behave in exactly the opposite way. While EX shows
entrainment in cooperative dialogs and disentrainment in competitive
dialogs, for IN it is the other way round. This is also well reflected
in Table \ref{tab:da_coop} showing the results of the t-tests for all
dialog acts, feature sets and entrainment measures in the cooperative
and competitive dialogs, respectively. Significant distance
differences indicating entrainment are marked by a + sign, significant
differences for disentrainment by a -- sign. Not significant cases are
marked by a zero. Profiles and tables show clear differences in
entrainment behavior in cooperative and competitive dialogs. Overall,
for dialog acts a lower number of significant entrainment cases is
observed in cooperative than in competitive dialogs (28 against 36\%
of all combinations between dialog acts and feature
sets). Furthermore, in cooperative dialogs disentrainment occurs more
often (14 against 9\%). A closer look at the single dialog acts
reveals that the supportive dialog acts EX, CL, and replies on average
undergo more entrainment and less disentrainment in cooperative
dialogs than in competitive ones, whereas for the not-supportive
dialog acts IN and questions, the pattern is the opposite.

\begin{table}
\begin{scriptsize}
\begin{tabular}{lllllllllllll}
\hline
\multicolumn{13}{l}{Cooperative dialogs} \\
\hline
set & AC & AL & CH & CL & EX & IN & QW & QY & RE & RN & RW & RY \\
\hline
GEN & + 0 & + -- & + + & + 0 & + 0 & 0 0 & +  0 & + + & + 0 & 0 0 & + 0 & 0 0 \\
GF0 & -- + & + 0 & 0 0 & 0 0 & 0 -- & 0 -- & -- -- & 0 0 & + 0 & 0 0 & 0 0 & -- 0 \\
IP & -- -- & 0 0 & 0 0 & + 0 & + + & -- -- & 0 0 & + + & 0 0 & + + & + + & 0 0 \\
ACC & 0 0 & 0 0 & 0 0 & 0 0 & 0 0 & -- -- & -- -- & 0 0 & + + & + + & 0 0 & 0 0 \\
RHY & 0 -- & 0 0 & + 0 & + 0 & + 0 & 0 0 & 0 -- & 0 0 & + 0 & + 0 & + 0 & + 0 \\
\hline
\multicolumn{13}{l}{Competitive dialogs} \\
\hline
set & AC & & CH & CL & EX & IN & QW & QY & RE & RN & RW & RY \\
\hline
GEN & 0 0 & & + 0 & 0 0 & + 0 & 0 0 & + 0 & + 0 & + 0 & 0 0 & + 0 & 0 0 \\
GF0 & + 0 & & + 0 & 0 0 & + + & + + & + 0 & + 0 & + + & + 0 & + + & + 0 \\
IP & 0 0 & & + -- & + + & -- -- & + + & + + & + 0 & + + & 0 -- & + 0 & + + \\
ACC & 0 0 & & 0 0 & -- -- & -- -- & 0 0 & 0 0 & 0 0 & 0 0 & -- -- & 0 0 & 0 0 \\
RHY & 0 0 & & 0 0 & 0 0 & + 0 & + + & 0 0 & + + & + + & 0 0 & 0 0 & 0 0 \\
\end{tabular}
\end{scriptsize}
\caption{Entrainment by dialog act, feature set, and distance measure
  in cooperative (upper half) and competitive (lower half) dialogs. Each cell shows a dialog act -- feature set
  pairing. Convergence is shown on the cell's left side, synchrony on
  the right side. + indicates entrainment, -- disentrainment, and 0 not
  significant. Significance is based on t-tests comparing within- and across-dialog distance measures. Remarkably, no AL act was found in the
  competitive data.}
\label{tab:da_coop}
\end{table}

\subsection{Entrainment by dialog act grouping}

\paragraph{Authority and support}

Tables \ref{tab:ng_coop} and \ref{tab:ng_comp} show the entrainment
and disentrainment proportions separately for cooperative and
competitive dialogs and for dialog act authority and
supportiveness. In cooperative dialogs high-authority dialog acts show
more entrainment and less disentrainment than low-authority ones. In
competitive dialogs the ratio is nearly balanced. For supportiveness
there is only a difference to report for the competitive dialogs:
supportive dialog acts show much less entrainment than non-supportive
ones.

Table \ref{tab:as} captures the interaction of authority and support
by showing the proportions of dialog acts exhibiting entrainment (+)
and disentrainment (--) for each pairing of authority and support in
cooperative and competitive dialogs. In cooperative dialogs the
highest entrainment values are shown for supportive high authority
acts ($0.34$) and not supportive low authority acts ($0.33$). A high
amount of disentrainment is given for not supportive high authority
acts ($0.5$). For competitive dialogs most entrainment is found for
the not supportive acts ($0.6$ and $0.47$ for high and low authority,
respectively).

\begin{table}
\begin{center}
\begin{tabular}{l|ll|ll|ll|ll|}
\cline{2-9}
{} &   \multicolumn{2}{l|}{authority} & \multicolumn{2}{l|}{support} & \multicolumn{2}{l|}{frequency} & \multicolumn{2}{l|}{predictability} \\
\cline{2-9}
{}       & +   & --  & +   & --  & +   & --  & +   & -- \\
\hline
\multicolumn{1}{|l|}{high/yes} & 0.5 & 0.1 & 0.3 & 0.1 & 0.4 & 0.1 & 0.3 & 0.1  \\
\multicolumn{1}{|l|}{low/no}   & 0.2 & 0.3 & 0.3 & 0.1 & 0.1 & 0.1 & 0.2 & 0.1 \\
\hline
\end{tabular}
\end{center}
\caption{Proportion of dialog acts exhibiting entrainment (+) and disentrainment (--) for
  authority, support, frequency and local predictability levels in
  cooperative dialogs. Proportions are calculated within each level,
  which implies that the remainder proportion (of neither entrainment
  nor disentrainment) is 1 minus the proportion for + and --,
  e.g. $1-0.5-0.1=0.4$ for high authority.}
\label{tab:ng_coop}
\end{table}

\begin{table}
\begin{center}
\begin{tabular}{l|ll|ll|ll|ll|}
\cline{2-9}
{} &   \multicolumn{2}{l|}{authority} & \multicolumn{2}{l|}{support} & \multicolumn{2}{l|}{frequency} & \multicolumn{2}{l|}{predictability} \\
\cline{2-9}
{} & + & -- & + & -- & + & -- & + & -- \\
\hline
\multicolumn{1}{|l|}{high/yes} & 0.5 & 0.1 & 0.2 & 0.2 & 0.6 & 0.0 & 0.6 & 0.0  \\
\multicolumn{1}{|l|}{low/no}   & 0.6 & 0.0 & 0.7 & 0.0 & 0.5 & 0.0 & 0.2 & 0.3 \\
\hline
\end{tabular}
\end{center}
\caption{Proportion of dialog acts exhibiting entrainment (+) and disentrainment (--)  for
  authority, support, frequency and local predictability levels in
  competitive dialogs. Proportions are calculated within each level,
  which implies that the remainder proportion (of neither entrainment
  nor disentrainment) is 1 minus the proportion for + and --,
  e.g. $1-0.5-0.1=0.4$ for high authority.}
\label{tab:ng_comp}
\end{table}

\begin{table}
\begin{center}
\begin{tabular}{llllll|llll}
\cline{3-10}
\multicolumn{2}{}{} & \multicolumn{4}{l|}{cooperative} & \multicolumn{4}{l}{competitive} \\
\cline{3-10}
& authority & \multicolumn{2}{l}{high} &  \multicolumn{2}{l|}{low} & \multicolumn{2}{l}{high} & \multicolumn{2}{l}{low} \\
\hline
\multicolumn{2}{}{} & + & -- & + & -- & + & -- & + & -- \\
\cline{3-10}
support & yes & 0.34 &  0.04 &  0.2&  0.25 & 0.28&  0.18 &   0.1&  0.0 \\
& no  &  0.0 &  0.5 & 0.33&  0.12 &   0.6&  0.0 & 0.47&  0.03 \\
\hline
\end{tabular}
\end{center}
\caption{Entrainment (+) and disentrainment (--) probabilities for all
  authority (columns: high, low) and support (rows: no, yes) level
  combinations in cooperative and competitive dialogs. Probabilities
  are calculated within each level combination, which implies that the
  probability of neither entrainment nor disentrainment is 1 minus the
  probabilities for + and --.}
\label{tab:as}
\end{table}

The impact of the factors authority and support was further tested by
a linear mixed effects model with random slopes with distance delta
$d$ as the dependent variable (values below 0 indicate entrainment),
dialog condition (cooperative vs. competitive), authority (high
vs. low), support (yes vs. no) as fixed effects, and speaker as a
random effect. The test reveals a significant impact of dialog
condition ($\chi^2=22.8944$, $p<0.0001$): $d$ is smaller in
competitive than in cooperative dialogs. Further significant
interactions are observed for all effect combinations ($\chi^2 \geq
27.1360$, $p<0.0001$), i.e. authority and supportiveness interact, and
both as well as their interaction behave differently in cooperative
and competitive dialogs. We further tested this behavior by
re-applying linear mixed effects random slope models for authority and
support as fixed effects separately for the cooperative and for the
competitive subset.

For the competitive dialogs no further significant difference was
found -- neither for authority ($\chi^2=0.0013$, $p=0.9712$) nor for
support ($\chi^2=0.0022$, $p=0.9122$) nor for their interaction
($\chi^2=2.6425$, $p=0.1040$). For cooperative dialogs data alone,
there were no significant effects for authority ($\chi^2=0.3951$,
$p=0.5296$) or support ($\chi^2=0$, $p=0.9974$), but their interaction
was significant ($\chi^2=25.1918$, $p<0.0001$).  Further splitting the
cooperative dialog data by authority and re-applying a linear mixed
effects model with support as fixed effect shows no significant
difference for the low-authority dialog acts ($\chi^2=0.0245$,
$p=0.8757$), but a significant difference for the high authority ones
($\chi^2=5.3074$, $p=0.02124$): high authority dialog acts show more
entrainment if they are also supportive.

\begin{figure}
  \begin{center}
    \includegraphics[width=12cm]{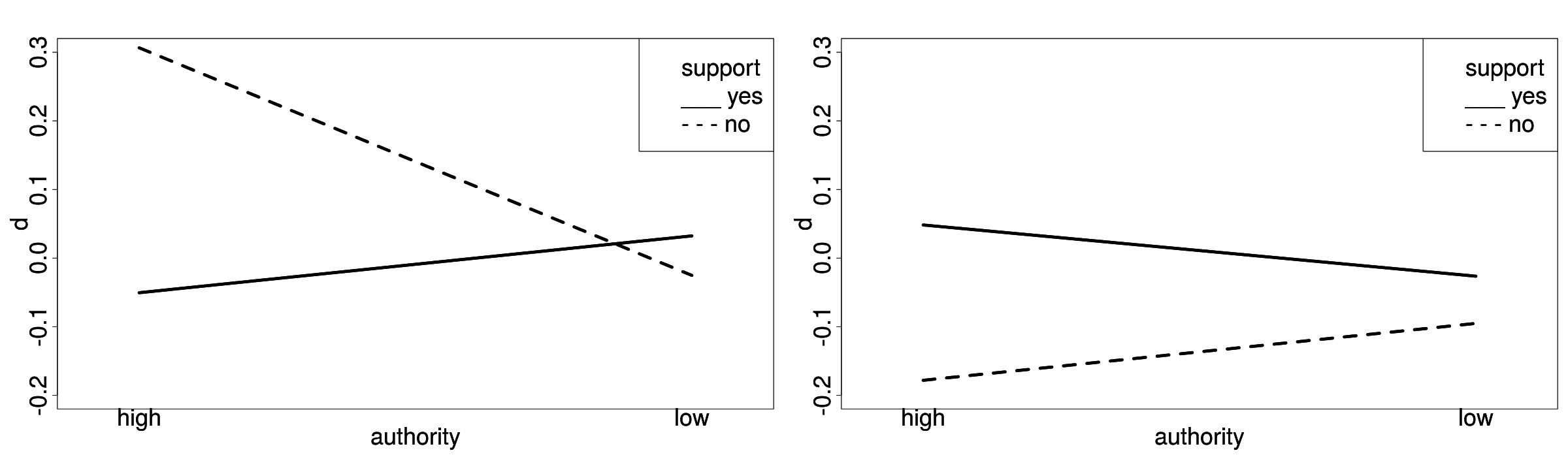}
    \caption{Interaction plot for authority and support in cooperative
      (left) and competitive (right) dialogs.}
    \label{fig:interact_as}
  \end{center}
\end{figure}

This interplay of authority and support in cooperative and competitive
dialogs is shown in Figure \ref{fig:interact_as}. Though only
significant for cooperative dialogs an opposite trend for
high-authority dialog acts is visible in these interaction plots. In
cooperative dialogs high-authority dialog acts show more entrainment
if they are also supportive, whereas in competitive dialogs there is
an opposite tendency: for these dialogs not supportive high-authority
dialog acts show most entrainment.

\paragraph{Frequency and local predictability}

In Tables \ref{tab:ng_coop} and \ref{tab:ng_comp} the general trend is
visible that high predictability and high frequency are related to a
higher amount of entrainment. Disentrainment is raised for
unpredictable dialog acts in the competitive setting only.

\begin{figure}
  \begin{center}
    \includegraphics[width=12cm]{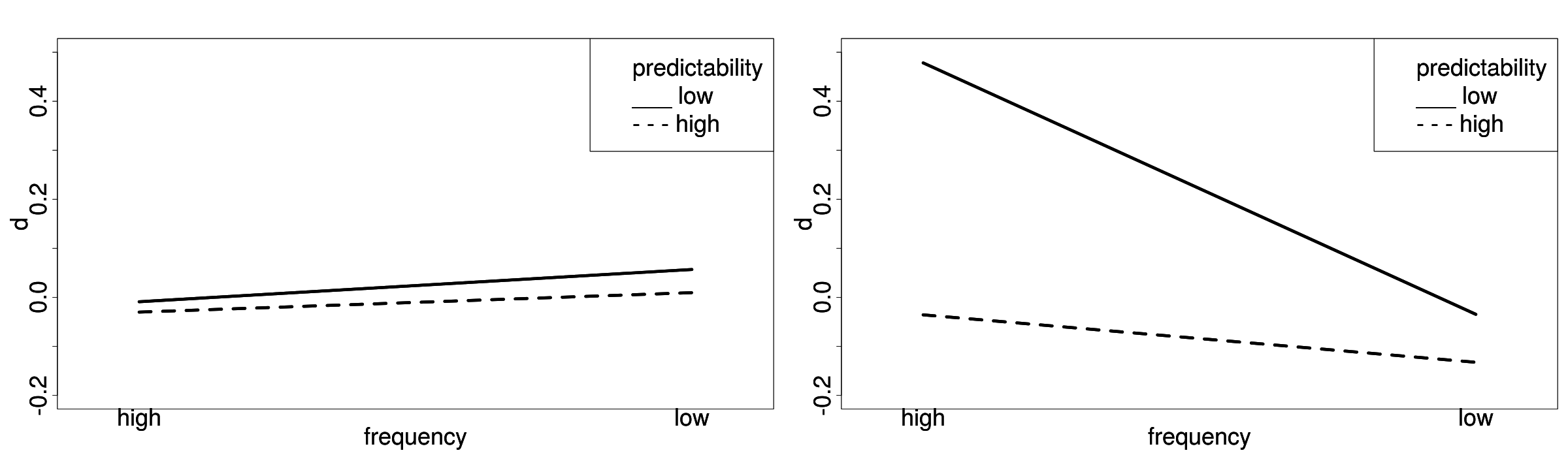}
    \caption{Non-interaction plot for frequency and local
      predictability in cooperative (left) and competitive (right)
      dialogs.}
    \label{fig:interact_ng}
  \end{center}
\end{figure}

The impact of the factors frequency and local predictability was again
tested by a linear mixed effects random slope model with distance
delta $d$ as the dependent variable, dialog condition (cooperative
vs. competitive), frequency (high vs. low), local predictability (high
vs. low) as the fixed effects, and speaker as a random effect. Next to
the significant impact of dialog condition the test revealed
significant interactions for all effect combinations ($\chi^2 \geq
41.0847$, $p<0.0001$), i.e. frequency and local predictability
interact, and both as well as their interaction behave differently in
cooperative and competitive dialogs. We further tested this behavior
by re-applying linear mixed effects random slope models for frequency
and predictability as fixed effects separately for the cooperative and
for the competitive subset. For the cooperative subset we found a
weakly significant impact of frequency on $d$ in the expected
direction, i.e. more entrainment for high frequency ($\chi^2=2.7097$,
$p=0.0997$). Predictability did not have a significant impact
($\chi^2=0.2609$, $p=0.6095$). For the competitive subset we found a
further significant interaction ($\chi^2=11.0077$, $p=0.0009$) which
was due to a close to weakly significant impact of predictability
($\chi^2=2.6437$, $p=0.104$) for the high-frequency condition
only. Again this impact goes into the expected direction, i.e. less
entrainment for low predictability. These tendencies, though weak, are
further illustrated by the interaction plots in Figure
\ref{fig:interact_ng}: in cooperative dialogs more entrainment is
found at high frequency (w.s.), for competitive dialogs less
entrainment for low local predictability (n.s.).

\section{Discussion and conclusion}

\subsection{Dialog acts as entrainment units}

To the best of our knowledge this is the first study examining the
prosodic entrainment related to dialog acts in detail.

We offer two arguments why dialog acts might be a more appropriate
unit for entrainment measurement than e.g. adjacent turns: First,
dialog acts determine the value range of several acoustic
parameters. As an example from the {\em frequency code} paradigm
\cite{Ohala1994}, questions tend to be uttered with higher pitch than
answers. That's also why prosodic features can be successfully used
for automatic dialog act classification \cite{Shriberg1998,
  MR_ESSV2016}. Therefore, if one simply measures the similarity of an
acoustic feature between adjacent turns and not between dialog acts,
thus neglecting dialog act intrinsic value ranges, the entrainment
results run the risk of being obscured.

Second, cooperative speakers often simply cannot entrain in adjacent
turns on several variables for e.g. syntactic or pitch contour
patterns. To give an illustrative example: if a speaker asks a
question, a cooperative dialog partner will neither exactly repeat
this question, thus imitate it as a whole, nor will she imitate the
intonation pattern of the question, unless she wants to mock the
inquirer. A cooperative partner will instead stick to the conventions
for coherent DA sequencing \cite{Schegloff2006} and thus will give an
answer or ask back for clarification choosing appropriate intonation
patterns. In other words, if she imitates at all, she cannot imitate
the preceding turn but a reaction to this turn, which is an answer
dialog act which occurred earlier in the dialog.

\subsection{Dialog condition related entrainment differences}

Remarkably, in the current study we found overall more entrainment in
competitive than in cooperative dialogs which on first sight is in
contrast to our previous findings on the same data for sequence
comparisons of prosodic events \cite{RC_PP16}. There we found a
greater event sequence similarity in cooperative dialogs. However,
this finding was already put into perspective by another study on this
corpus \cite{Cole2016} comparing entrainment in cooperative and
competitive dialogs separately for several prosodic variables. These
variables showed highly varied patterns of entrainment and
disentrainment behavior in cooperative and competitive dialogs. In the
current study instead of focusing on few (sequence similarity)
variables we were operating with much larger feature sets which in
total we expect to give a more robust estimate about the overall
amount of entrainment. This robustness claim is further supported in
that we obtained the same tendencies for several random dialog act
segment pairings. In any case, as will become clear in the subsequent
parts of the discussion, in our data cooperative and competitive
behavior cannot simply be described by an overall entrainment
comparison, but needs a more fine-grained analysis based on intrinsic
characteristics of dialog acts.

\subsection{Dialog act related entrainment differences}

\paragraph{Game structuring}
Overall, game structuring, which is mainly carried out by the dialog
act RE undergoes entrainment to a large extent (cf. Table
\ref{tab:da_coop}). Thus interlocutors highly entrain on the level of
organizing the sequence of actions in the game.

\paragraph{Expected dialog act effects}
Based on the relations between entrainment and authority reported in
section \ref{sec:intro} we expected more entrainment in low authority
dialog acts than in high-authority ones (H1). Furthermore, based on
the potentially supportive nature of both entrainment and
disentrainment, we expected more entrainment in supportive (H2) and in
frequent dialog acts (H3), and disentrainment in unpredictable dialog
acts (H4). H3 is motivated by the enhancement of turn transition
smoothness by entrainment, and H4 by the support of marking unexpected
events. Finally, H2, H3, and H4 are expected to be more strongly
confirmed in the cooperative dialogs than in the competitive ones.

\paragraph{Authority and supportiveness}
From our data hypothesis H1 needs to be rejected. We did not find more
entrainment in low-authority dialog acts. Neither can H2 be confirmed
as is, since the pattern we found is more complex. From the
interaction of authority and supportiveness shown in Figure
\ref{fig:interact_as} and Table \ref{tab:as} one can conclude the
following: in cooperative dialogs those dialog acts are entraining
that are both of high authority and high supportiveness, as EX and CL,
whereas the not supportive IN strongly disentrains (cf. Table
\ref{tab:da_coop}). In competitive dialogs there is more entrainment
also for IN, the high-authority dialog act that is not supportive but
imposes an obligation to the interlocutor (cf. Table
\ref{tab:da_coop}).

From this one can conclude that at least in cooperative dialogs there
is a clearly distinctive entrainment behavior for different types of
authority. While -- in line with the literature -- executive authority
provokes disentrainment, supporting authority by knowledge shows
entrainment. The latter type of authority enables the speaker to
provide the information needed by the interlocutor to successfully
solve the game, which is thus further supported by accommodation.

\paragraph{Frequency and local predictability}
We found a weak frequency effect on entrainment in cooperative dialogs
only, so that H3 was partly confirmed. There frequent dialog acts
entrain more. In the context of conversation facilitation
\cite{Levitan2012, Nenkova2008} this frequency effect can be
interpreted to contribute to the joint cooperative effort to quickly
reach the common goal by smoothing the transitions from and to often
occurring building blocks in the dialog.

Local dialog act predictability did not have a significant effect on
entrainment, thus H4 is to be rejected. In our data we did not find a
sufficiently strong indication that a lack of predictability would be
the reason for cooperative disentrainment \cite{Perez2016}.

\paragraph{Active control in cooperative behavior}
Since we found entrainment differences among dialog acts and between
cooperative and competitive settings, it can be concluded that
entrainment is not only an automatic process in terms of a
perception-action loop. Rather it is also actively controlled in order
to provide support in joint cooperative actions. In \cite{RPNC_IS2015}
we found, for the same corpus, text-based differences between
cooperative and competitive dialogs related to word n-gram entropies
and proportions of pronouns, affirmations, as well as to selectional
preferences for dialog acts. Both could be well interpreted in terms
of the Gricean cooperative principle \cite{Grice1975} which consists
of the four conversation maxims of appropriate quantity, quality,
relevance, and manner, and in terms of Relevance Theory
\cite{Sperber1986, Sperber2004}.  Relevance Theory states that the
relevance of an utterance for the hearer is defined as a function of
positive cognitive effect and processing effort. The positive
cognitive effect reflects the importance of the conveyed information
for the hearer.  The processing effort is the needed labor for the
hearer to extract and make use of a conveyed information. Related to
communication behavior, a cooperative speaker is expected to maximize
the relevance in terms of providing important information in an
easy-to-process way.  How does this relate to our findings? In
addition to selectional preferences of dialog acts found in
\cite{RPNC_IS2015}, e.g. a preference of information-conveying dialog
acts like EX and CL in cooperative dialogs, in the current study we
also found different entrainment behavior for these dialog
acts. Notably, for EX and CL with the highest information content only
a single feature set undergoes disentrainment in cooperative dialogs
(GF0 for EX), while there are six instances of disentrainment in
competitive dialogs. Especially, the feature set ACC related to pitch
accents and thus to the encoding of information status
\cite{Pierrehumbert1990} disentrains for both dialog acts, which might
be used by the speakers to impede the processing of important new
information in the competitive setting.

Taken together the positive impact of authority by knowledge,
frequency, and information transmission on entrainment provide
evidence that entrainment is partly under active control to fulfill
the fourth Gricean maxim of manner, i.e. to appropriately convey
information, and -- in a relevance-theoretic sense -- to minimize the
processing effort for the interlocutor.

\subsection{Conclusion}
We measured entrainment in cooperative and competitive dialogs
separately for several dialog acts and for a large amount of
intonation, energy, and rhythmic variables derived from a
computational prosodic stylization. Overall, the speakers highly
entrain in dialog events serving to structure the game. For
cooperative dialogs we found more entrainment in frequent dialog acts
which can contribute to a smooth processing of frequently occurring
dialog units. Furthermore, it turned out that the concept of authority
as a source of entrainment needs to be subdivided into authority by
knowledge and executive authority, the former leading to entrainment
the latter to disentrainment in cooperative dialogs. Finally, the
finding that entrainment patterns differ as a function of dialog act
and dialog condition provides evidence that entrainment is not an
entirely automatic process but is at least in part actively controlled
as a component of voluntary cooperative or competitive behavior.

\section{Acknowledgments}
The work of the first author was financed by a grant of the Alexander
von Humboldt Foundation. The experiment design and data collection
phase of this project was funded by a grant from the
VolskwagenStiftung awarded to the third author as part of the Prosodic
and Gestural Entrainment project.

\bibliographystyle{gerabbrv}
\bibliography{RMC}

\end{document}